\documentclass[conference]{IEEEtran}
\IEEEoverridecommandlockouts
\usepackage{amstext} 
\usepackage{array}   
\newcolumntype{L}{>{$}l<{$}} 
\usepackage{cite}
\usepackage{amsmath,amssymb,amsfonts}
\usepackage{algorithmic}
\usepackage[linesnumbered,ruled]{algorithm2e}
\usepackage{graphicx}
\usepackage{textcomp}
\usepackage{xcolor}
\usepackage{array}
\usepackage{wrapfig}
\usepackage{multirow}
\usepackage{tabu}
\usepackage{url}
\usepackage{subcaption}

\def\BibTeX{{\rm B\kern-.05em{\sc i\kern-.025em b}\kern-.08em
    T\kern-.1667em\lower.7ex\hbox{E}\kern-.125emX}}
    
\newif\ifdraft
\draftfalse
\ifdraft
 \newcommand{\eunsungnote}[1]{ {\textcolor{red} { ***Eunsung: #1 }}}
 \newcommand{\shahzadnote}[1]{ {\textcolor{blue} { ***Shahzad: #1 }}}
\else
 \newcommand{\eunsungnote}[1]{}
 \newcommand{\shahzadnote}[1]{}
\fi
\newcommand{\ignore}[1]{}

\begin{document}

\title{Factorial Convolution Neural Networks}

\author{\IEEEauthorblockN{Jaemo~Sung}
\IEEEauthorblockA{\textit{DesignedAI, Korea} \\
jaemo.sung@designedai.com}
\and
\IEEEauthorblockN{Eun-Sung Jung}
\IEEEauthorblockA{\textit{Department of Software and Communication Engineering} \\
\textit{Hongik University, Korea}\\
ejung@hongik.ac.kr}
}

\maketitle

\begin{abstract}
In recent years, GoogleNet has garnered substantial attention as one of the base convolutional neural networks (CNNs) to extract visual features for object detection. However, it experiences challenges of contaminated deep features when concatenating elements with different properties. Also, since GoogleNet is not an entirely lightweight CNN, it still has many execution overheads to apply to a resource-starved application domain. Therefore, a new CNNs, FactorNet, has been proposed to overcome these functional challenges. The FactorNet CNN is composed of multiple independent sub CNNs to encode different aspects of the deep visual features and has far fewer execution overheads in terms of weight parameters and floating-point operations. Incorporating FactorNet into the Faster-RCNN framework proved that FactorNet gives \ignore{a 5\%} better accuracy at a minimum and produces additional speedup over GoolgleNet throughout the KITTI object detection benchmark data set in a real-time object detection system.
\end{abstract}

\section{Introduction}\label{sec:intro}
The past few years, has seen deep convolutional neural network (CNN)~\cite{LeCunBoserDenkerEtAl89, krizhevsky2012imagenet} as a standard feature extraction method for visual understanding tasks (e.g., object detection/classification/segmentation, image compression, image translation, and scene understanding~\cite{Rawat2017DeepCN, Ajmal2018ConvolutionalNN, Dhillon2019ConvolutionalNN}). Their excellent capability to encode learnable visual information from data results in an unprecedented performance ~\cite{russakovsky_imagenet_2015, Li2021ASO}. Moreover, tasks such as face recognition, autonomous driving, and intelligent medical treatment in the industry, which were once considered impossible, have been achieved using CNNs. 

VGG~\cite{szegedy_going_2015} and GoogleNet~\cite{liu_very_2015} have been one of successful CNN models that not only motivated other CNN models~\cite{howard2017mobilenets, 8578814} but also derived object detection methods such as Faster-RCNN~\cite{ren_faster_2017} and SSD~\cite{LiuAESRFB16}. They are motivated to model a big receptive field differently. However, to do so, VGG adopted very deep stacked convolution layers, each of which covers small receptive fields of a 3x3 kernel size. Again, the deep nature of the structure requires significant computation and heavy memory overhead, thus hindering its deployment on resource-restricted applications. Therefore, to overcome this drawback of VGG, GoogleNet and its variants~\cite{szegedy2015rethinking} employ a so-called inception layer composed of multiple convolution operations with different levels of kernel sizes. With this, GoogleNet reduces the depth of CNN while covering somewhat large receptive fields. However, its embedded concatenation operation within the inception layer can eventually clutter hidden features representing different granularities, making it hard to draw the long-depth structured visual information at the final CNN feature maps. Although GoogleNet may have a lower overhead than VGG, its heaviness prevents possible deployment in some applications involving high-resolution input images or resource-starved SOCs (System-On-Chips). Thus solving the challenges mentioned above of VGG and GoogleNet, a new CNN model, known as FactorNet, composed of multiple independent sub CNNs to encode the different aspects of deep visual features as well as reduce the execution overhead in terms of weight parameters and floating-point operations has been proposed. 

To achieve easy deployment of the CNN models on resource-restricted systems, attempts were made to lighten the models to reduce the model parameters and subsequent computational cost. MobileNet~\cite{howard2017mobilenets} and its variants~\cite{8578572} employed the depth-wise separable convolution layer, which decomposes the conventional convolution layer into depth-wise convolution operator and point-wise convolution operator. In addition, ShuffleNet~\cite{8578814} adapted point-wise group convolution and employed channel-shuffle to exchange the information among the groups. Thus, MobileNet and ShuffleNet can be viewed as a local factorization method for the CNN model, in which the convolution layer is locally decomposed into simpler ones. On the contrary, FactorNet utilizes a more global factorization over the CNN structure than the convolution layer, making it better to extract consistent deep spatial features while lightening the CNN model.

\eunsungnote{more detailed description including paper structure.}
This article is structured as follows. Section~\ref{sec:factornet} describes the details of the proposed method by comparing the architecture of FactorNet to that of GoogleNet. Section~\ref{sec:exp} presents the experimental results from applying GoogleNet and FactorNet to the practical object detection application in autonomous vehicles. Section~\ref{sec:conclusion} concludes with summary of contribution and experimental results.

\section{Proposed Method}\label{sec:factornet}
In this section, the deep neural networks are described in comparison to existing CNN models. 

\begin{figure*}
\centerline{\includegraphics[width=0.95\textwidth]{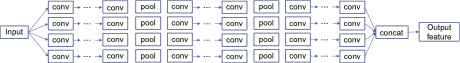}}
\caption{Example of FactorNet Architecture: conv and pool represent the convolution and pooling operations.\label{fig:factornet_archi}}
\end{figure*}
\subsection{Overview}
First, a pseudo-architecture of the proposed FactorNet comprising four factors as in Fig.~\ref{fig:factornet_archi} is presented. FactorNet usually is defined by several sub-CNN-networks referred to as factors. Each factor can be specified by different filter sizes but has a smaller output channel size than conventional CNNs as VGG16 or GoogleNet. The advantages of the proposed FactorNet is as follows:
\begin{itemize}
    \item FactorNet can encode multiple levels of independent deep spatial information. In practice, object detection/recognition tasks are based on natural scenes such as the autonomous driving environment, comprising many different objects varying in size, texture, and rigidness. It is not optimal for a single but large CNN to encode all such different object information appropriately. By dividing a single big CNN into multiple independent sub-CNNs, FactorNet, however, makes it possible for each factor to encode its inherent features in different granularity. Therefore, as a whole, it results in more independent features that deliver more clear information to the next work, such as object detection/classification.
    \item FactorNet is composed of sub-factors in the structural level, each of which generally comes with minor input and output channels than the conventional CNN. These minor inputs and outputs reduce the model parameters to be optimized and can alleviate the overfitting problem from which the conventional CNN suffers. Also, due to the fewer floating-point operations, its computational efficiency is higher compared GoogleNet. For such reasons, FactorNet's application suitability is extensive. 
    \item One essential aspect to consider when optimizing and deploying a sizable CNN-based model is its parallel computation ability to get maximum utilization of a given hardware resource. So far, it has not been easy to realize the model parallelism with a conventional CNN due to its layer-wise dependencies. However, the proposed FactorNet inherently implements the model parallelism because each sub-factor can be executed independently on a different computing node.
\end{itemize}

\begin{figure}[htb]
	\centering
	\includegraphics[width=0.9\columnwidth]{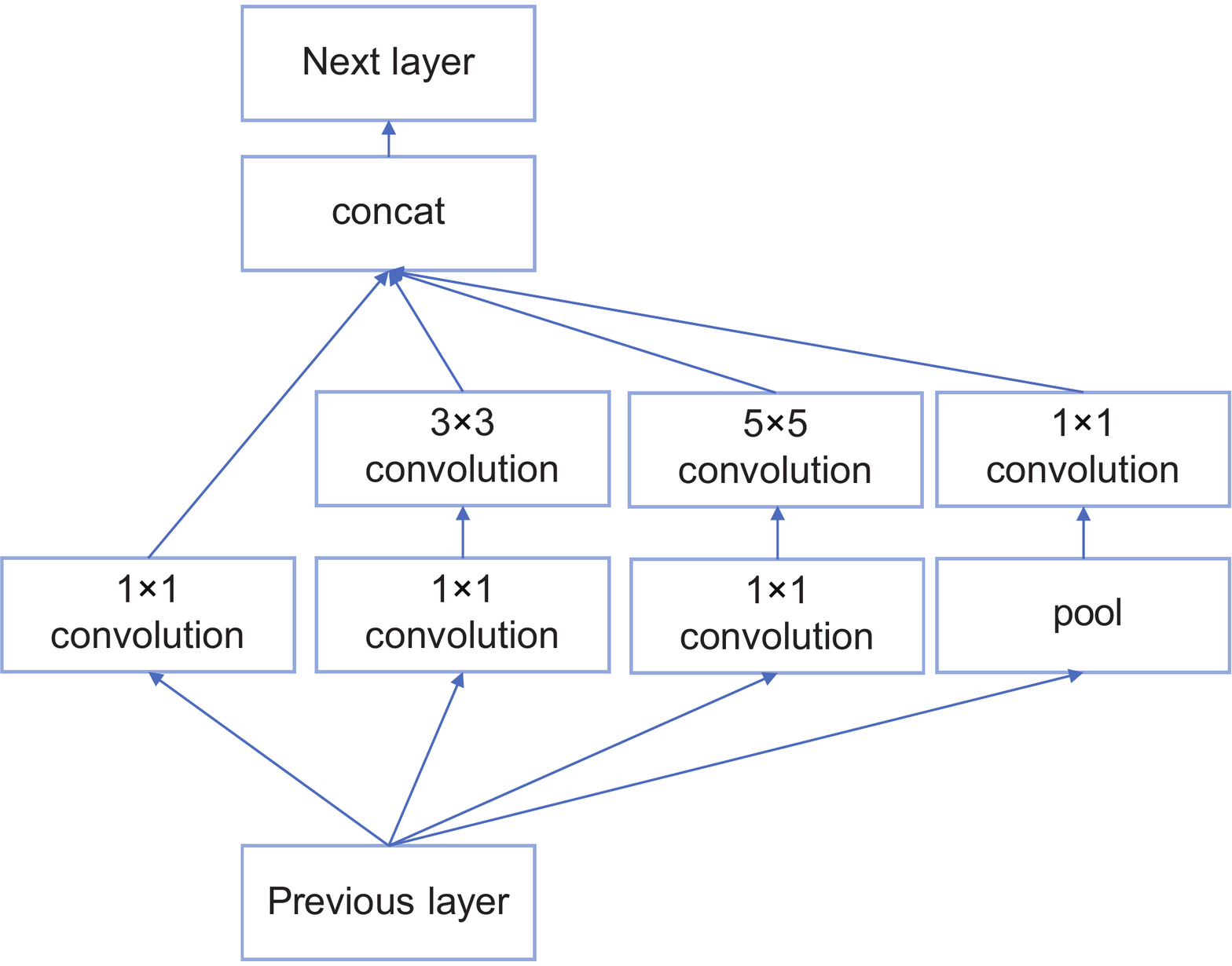}
	\caption{Internal architecture of inception layer of GoogleNet.}
	\label{fig:googlenet_archi}
\end{figure}

First, the inception layer and the core module of GoogleNet were presented in Fig.~\ref{fig:googlenet_archi} whereas the architectural specification of GoogleNet was also showcased in Table~\ref{tbl:googlenet_spec}, with which we compared to validate our proposed FactorNet. For brevity, the detailed specifications of each inception layer of GoogleNet were omitted. Interested readers can refer to the original paper~\cite{liu_very_2015}. In the case of 1920x1080x3 (width $\times$ height $\times$ channels) FHD image input, GoogleNets extract the 120x67x832 features up to 4e layer. Starting from the input, this is normally performed by visiting the layers downwards from la to 4a and executing the operations defined in each layer. We present the number of weights (i.e., parameters) and operations required for each layer of the GoogleNet in Table ~\ref{tbl:googlenet_spec}.

\subsection{Architecture of {GoogleNet}}

Specifically, the inception layers are an effective way to encode visual features from large receptive fields by exploiting different levels of kernel size rather than increasing the depth of layers. Yet, its concatenation operation in the inception layer can make it difficult to express the long-depth features encoded by multiple consecutive layers, thus increasing execution cost in terms of floating-point operation and memory storage. Therefore to resolve this problem, FactorNet was re-presented without the concatenation operation.

\begin{table}
	\centering
	\caption{Specification of GoogleNet for 1920x1080 FHD input image: the details of each inception layer are given in \cite{szegedy_going_2015}. (C: convolution, P: max pooling)}
	\includegraphics[width=0.95\columnwidth]{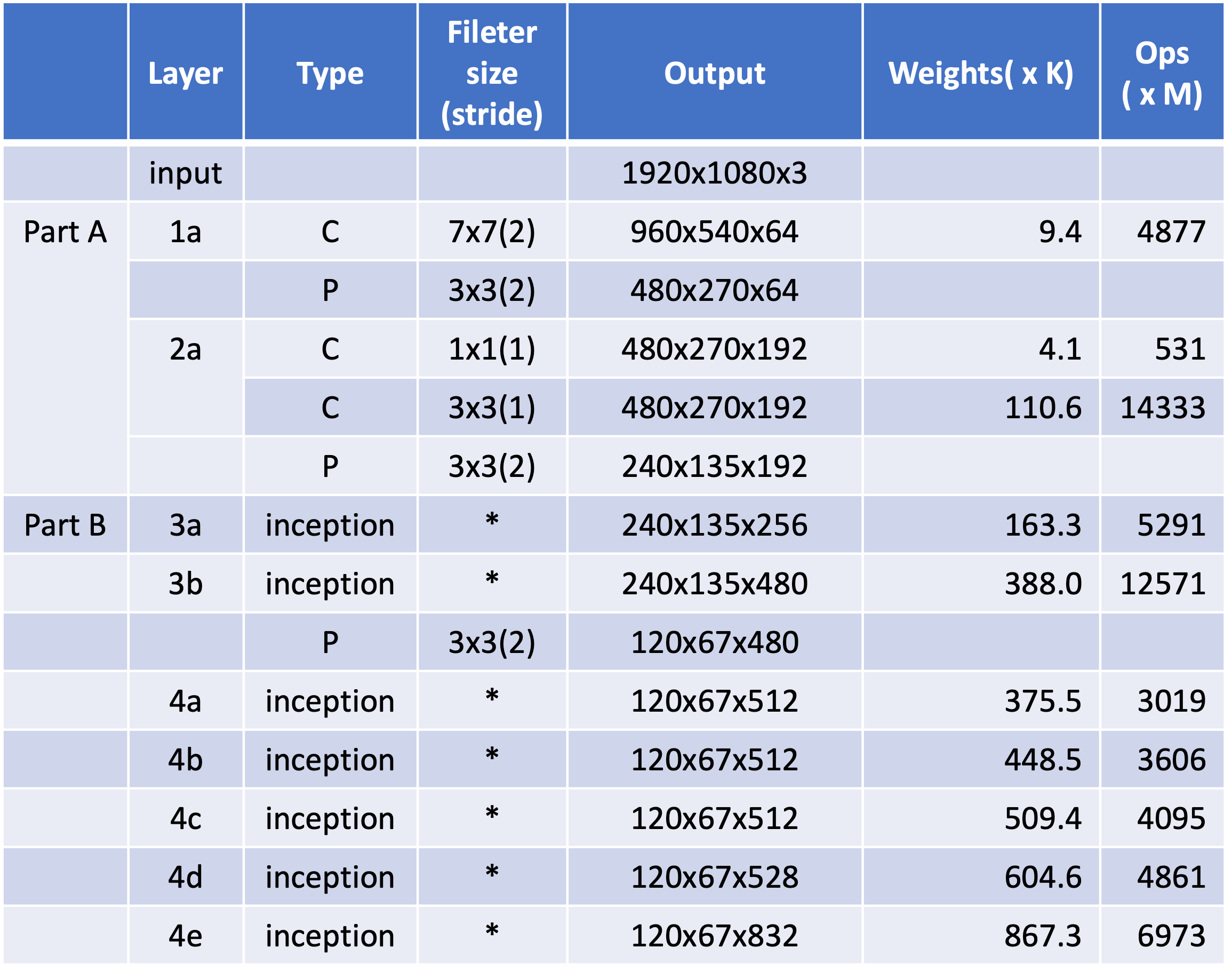}
	\label{tbl:googlenet_spec}
\end{table}

\subsection{FactorNet}
\begin{figure}[htbp]
	\centering
	\includegraphics[width=0.9\columnwidth]{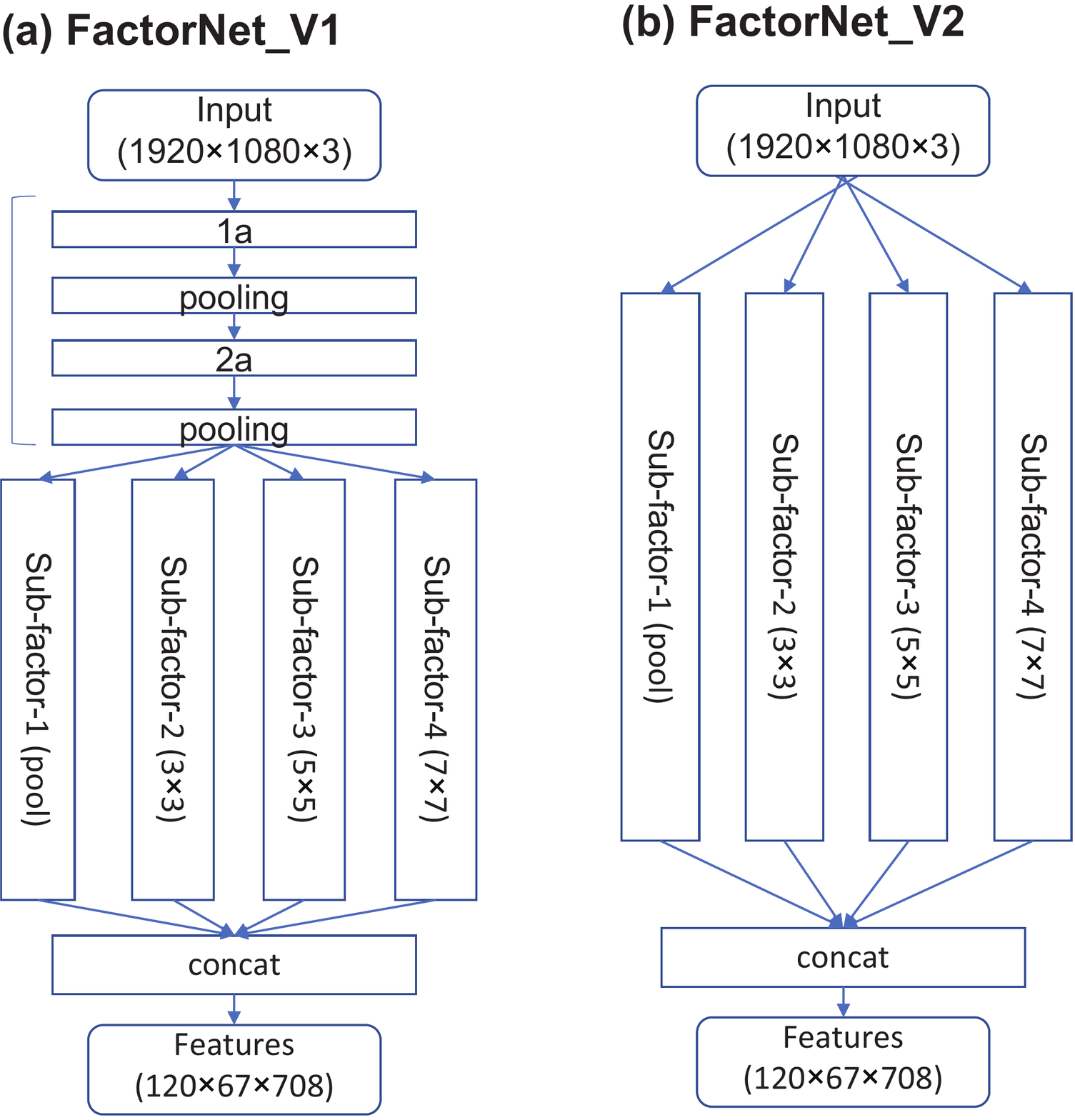}
	\caption{Architecture of the proposed FactorNets.}
	\label{fig:factornet_v1v2}
\end{figure}

\begin{figure}[htbp]
	\centering
	\includegraphics[width=0.9\columnwidth]{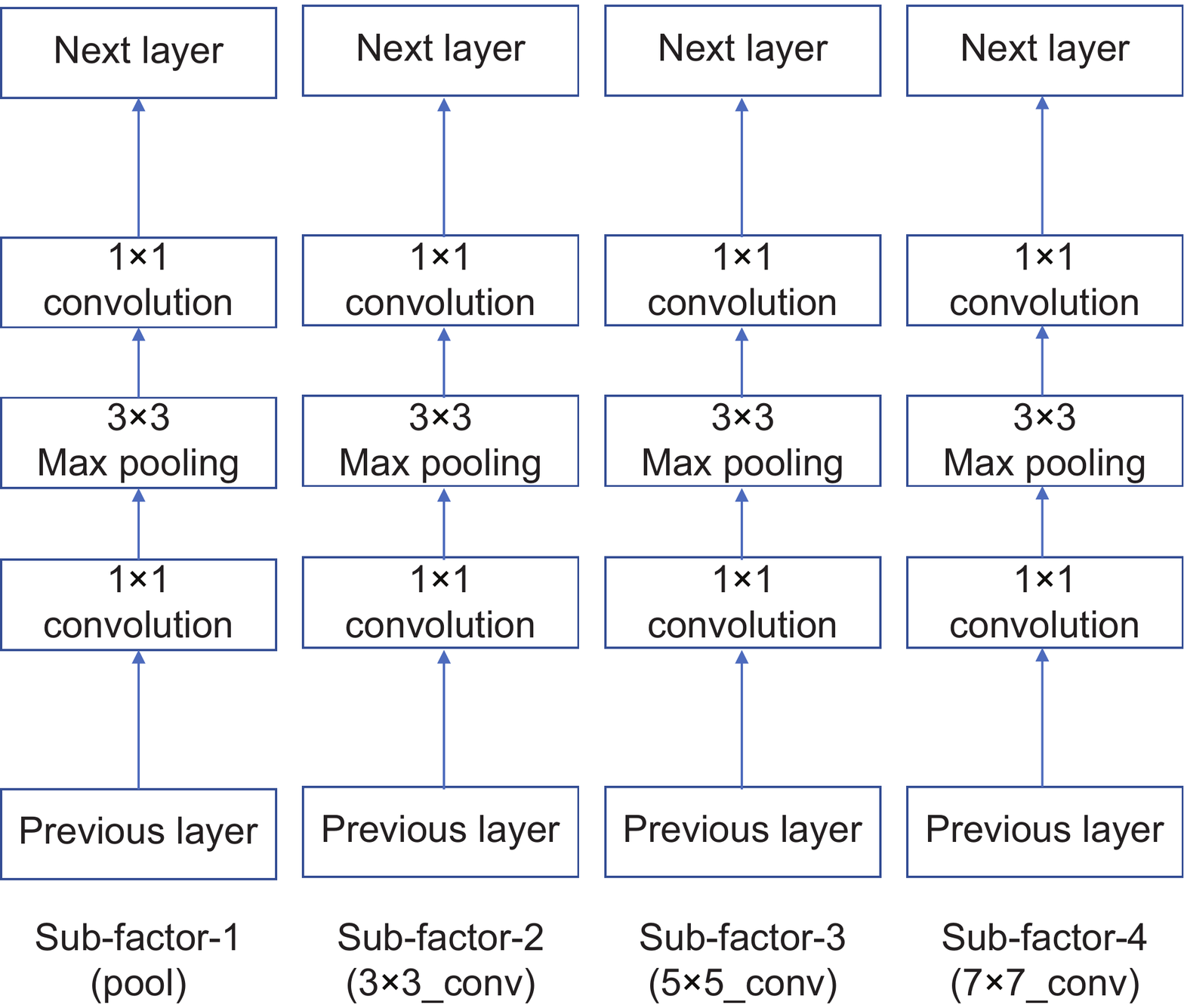}
	\caption{Sub-factors in FactorNet.}
	\label{fig:factornet_subfactors}
\end{figure}

In this paper, two types of FactorNets were designed. The first had the two initial layers of GoogleNet annotated by Part A in Table~\ref{tbl:googlenet_spec}, but the other layers are composed of four sub-factors. \eunsungnote{Part B is other than Part A?} This FactorNet was denoted by \emph{FactorNet\_V1} and showed its architecture in Fig.~\ref{fig:factornet_v1v2}(a). The other, denoted by \emph{FactorNet\_V2}, is fully defined by four sub-factor subnets from input to the final layer as shown in Fig.~\ref{fig:factornet_v1v2}(b). Each sub-factor is designed to encode different types of visual features, and the types of sub-factors that we commonly used for both FactorNet were presented\_V1 and FactorNet\_V2 in Fig.~\ref{fig:factornet_subfactors}. The first sub-factor is defined by 3$\times$3 max-pooling operation. However, other sub-factors conduct the spatial convolution with increasing filter size, that is, 3$\times$3, 5$\times$5 and 7$\times$7, which make it possible to represent the long-depth features encoded by different extents of receptive fields. We give the detailed architectural specifications of FactorNet\_V1 in Fig.~\ref{tbl:factornet_v1_spec} and FactorNet\_V2 in Fig.~\ref{tbl:factornet_v2_spec}.

\begin{figure}[htbp]
	\centering
	\includegraphics[width=0.9\columnwidth]{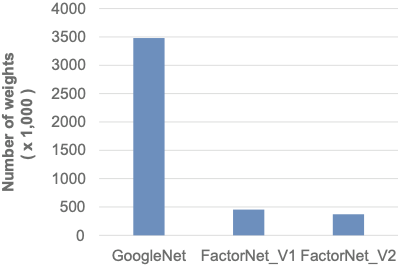}
	\caption{Comparison of the number of weights over CNN models.}
	\label{fig:comparison_weights_cnn_models}
\end{figure}

\begin{figure}[htbp]
\centerline{\includegraphics[width=0.9\columnwidth]{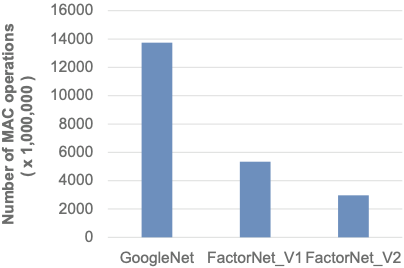}}
\caption{Comparison of the number of floating-point multiply-accumulate (MAC) operations for the forward propagation over CNN models.\label{fig:comparison_cnn_models}}
\end{figure}

One of the obvious advantages of such FactorNets is that they can compactly represent the model without losing the representation capability. The data forward in Fig.~\ref{fig:comparison_weights_cnn_models} and \ref{fig:comparison_cnn_models} was processed by comparing the model weights (parameters) and floating-point multiply-accumulate (MAC) operations. Thus, it is surprising that both FactorNets, FactorNet\_V1, and FactorNet\_V2, show significantly fewer overheads in weights and operations than GoogleNet. They require 7x fewer weights (i.e., the model parameters) to be stored in memory, as well as 3-4x fewer floating-point operations than GoogleNet. It was also noted that this model representation and computation efficiency are important when one wants to deploy a CNN model on the resource-restricted HW platform such as Smartphone, embedded device, SOC (System on chip), etc. 

Furthermore, the data representation capability of FactorNet does not decrease due to its compact model representation property. 
A comparison between the number of features of FactorNet and GoogleNet was shown in Fig. 7. FactorNet\_V1 produces slightly fewer features than GoogleNet. However, FactorNet\_V2 generates much richer features, which is approximately two times more than GoogleNet.

\begin{figure}[htbp]
	\centering
	\includegraphics[width=0.9\columnwidth]{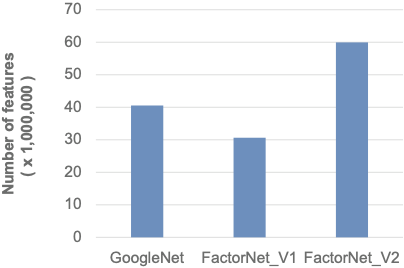}
	\caption{Comparison of the number of features over CNN models.}
	\label{fig:comparison_features_cnn_models}
\end{figure}

\begin{figure*}[htbp]
     \centering
     \includegraphics[width=0.8\textwidth]{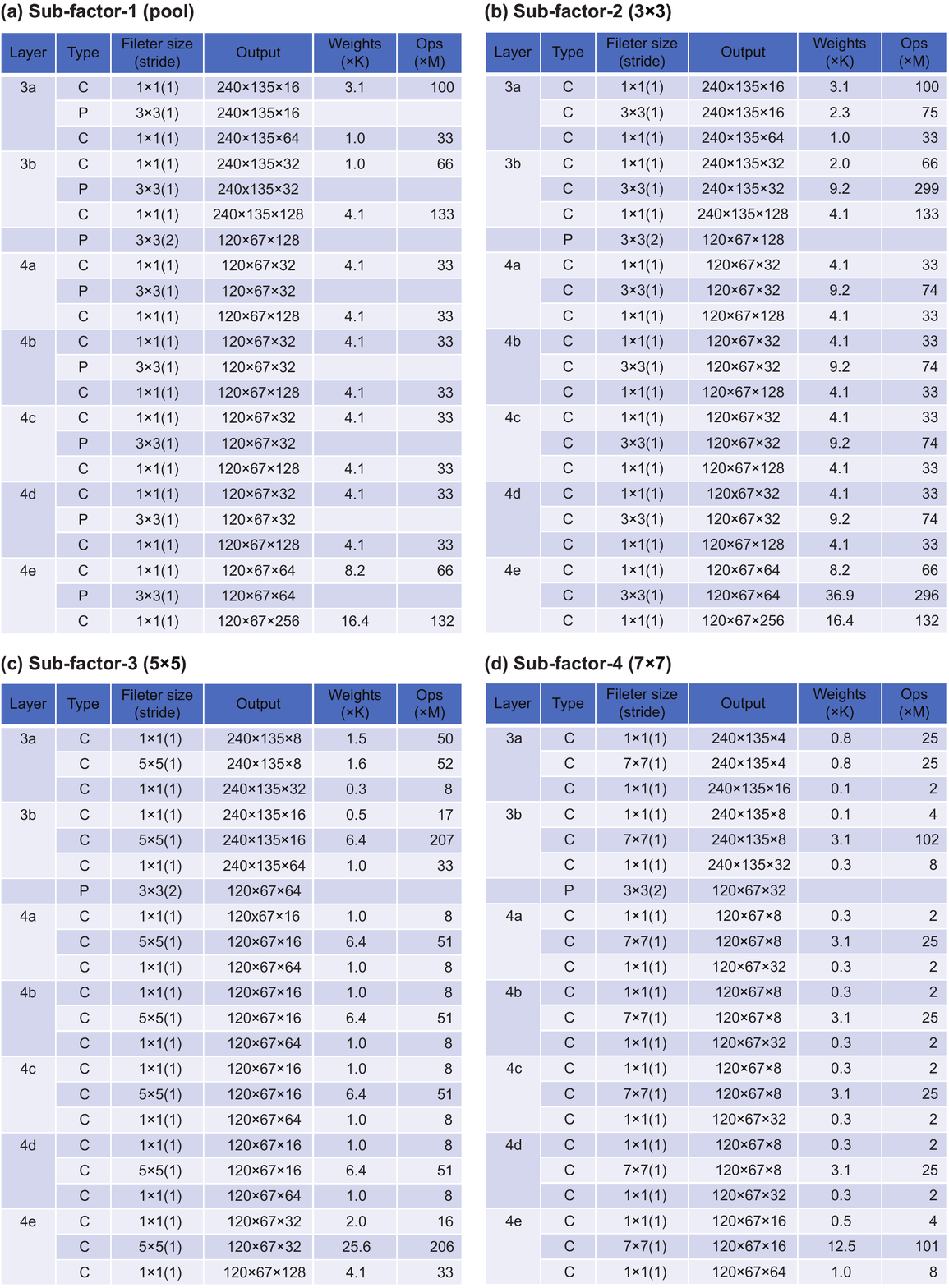}
     \caption{Specifications of sub-factors of FactorNet\_V1 in Fig.~\ref{fig:factornet_v1v2}(a) in the case of 1920x1080x3 inputs.}
     \label{tbl:factornet_v1_spec}
\end{figure*}

\begin{figure*}[htbp]
     \centering
     \includegraphics[width=0.8\textwidth]{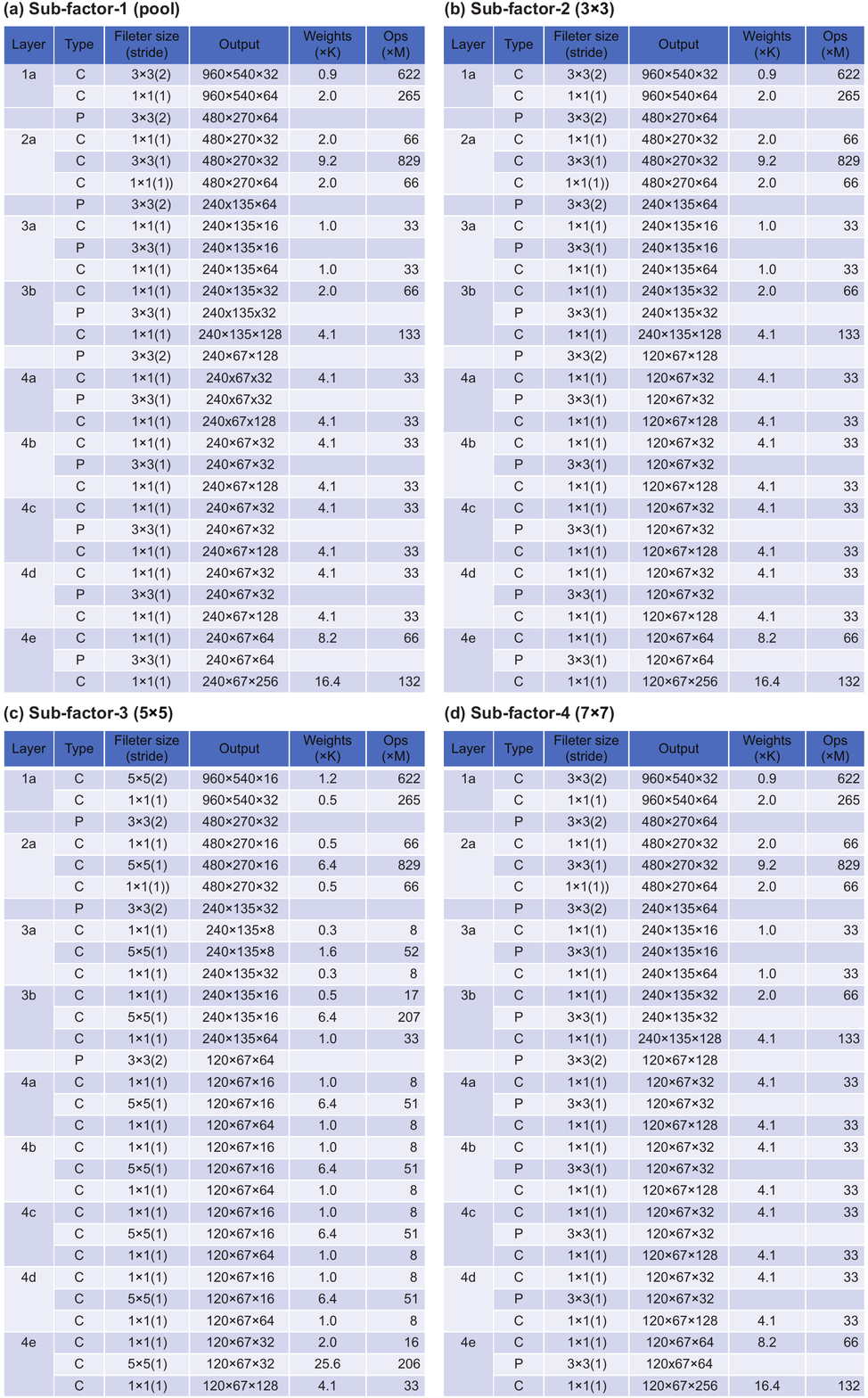}
     \caption{Specifications of sub-factors of FactorNet\_V2 in Fig.~\ref{fig:factornet_v1v2}(b) in the case of 1920x1080x3 inputs.}
     \label{tbl:factornet_v2_spec}
\end{figure*}

\section{Experimental Results}\label{sec:exp}

In evaluating the proposed FactorNet\_V1 and FactorNet\_V2, both were applied to the object detection task and juxtaposed to GoogleNet, which is a more expensive CNN model than FactorNets as mentioned in Section~\ref{sec:factornet}. Furthermore, a fair comparison was effected by subjecting both FactorNets and GoogleNet to experiments conducted in controlled conditions.

It was observed that object detection performance was generally affected by many other factors besides the performance of CNN features such as classification methods, parameter optimization methods, training data set size, input image augmentations, etc. Therefore, the maximization of the performance of CNN models for a particular data domain by optimizing its contributing factors is not the target of this study and is outside the scope of this work. 

First, FactorNet is tested alongside GoogleNet for car detection tasks by using KITTI benchmark dataset~\cite{geiger_are_2012} composed of approximately 1242 $\times$ 375 7481 images in the driving environment. This test was executed by applying both FactorNets and GoogleNet to the Faster-RCNN framework proposed in~\cite{ren_faster_2017}. In addition, we divided the dataset by 3712 images for training and 3769 images for testing. So, to measure the bias on pre-training, the model was trained from cold-start, i.e., random initial weight parameter values, and the pre-trained model based on ImageNet dataset~\cite{geiger_are_2012}.

\begin{figure*}[htbp]
     \centering
      \begin{subfigure}[b]{0.48\textwidth}
         \centering
         \includegraphics[scale=0.48]{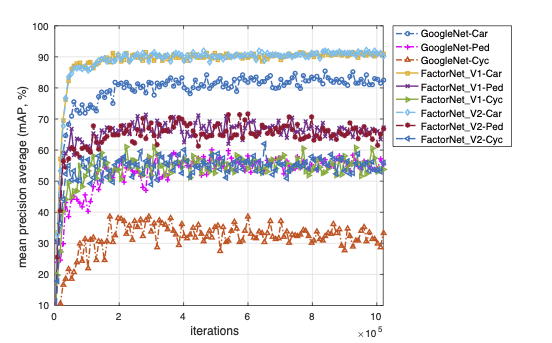}
         \caption{pre-trained model}
     \end{subfigure}
     \hfill
     \begin{subfigure}[b]{0.48\textwidth}
         \centering
         \includegraphics[scale=0.48]{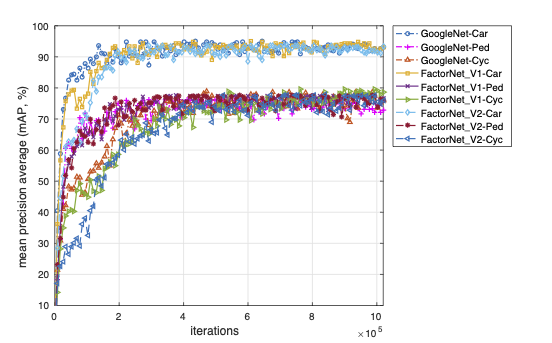}
         \caption{cold-started model}
     \end{subfigure}
       \caption{Mean Average Precision (mAP) for KITTI test data set over training iterations, when the models are trained from (a) pre-trained model and (b) cold-started model.}
	\label{fig:map_kitti}
\end{figure*}

\begin{table}[h]
    \begin{subtable}[h]{0.45\textwidth}
        \centering
        \begin{tabular}{|c | c| c| c|}
        \hline
            & Car & Pedestrian & Cyclist \\
        \hline \hline
        GoogleNet & 85.61\% & 59.94\% & 38.66\%\\
        FactorNet\_V1 & 91.42\% & 71.15\% & 61.07\%\\
        FactorNet\_V2 & \bf{92.4}\% & \bf{71.54}\% & \bf{62.67}\%\\
        \hline
       \end{tabular}
       \caption{mAP in the case of imagenet pre-trained model}
    \end{subtable}
    \newline
    \vspace*{1 cm}
    \newline
    \begin{subtable}[h]{0.45\textwidth}
        \centering
        \begin{tabular}{|c | c| c| c|}
        \hline
            & Car & Pedestrian & Cyclist \\
        \hline \hline
        GoogleNet & 94.82\% & 76.16\% & 78.72\%\\
        FactorNet\_V1 & \bf{94.94}\% & \bf{77.43}\% & \bf{79.56}\%\\
        FactorNet\_V2 & 93.71\% & 77.41\% & 77.67\%\\
        \hline
       \end{tabular}
        \caption{mAP in the case of cold-started model}
     \end{subtable}
     \caption{Comparison of detection performances in mAP for KITTI test set: results are based of easy object cases defined in~\cite{geiger_are_2012}}
     \label{tbl:detection_comparison}
\end{table}

Table 2 demonstrates the mean Average Precision (mAP) results of FactorNets for detecting car, pedestrian, cyclist class based on the KITTI test data set, compared to GoolgeNet. For all cases, FactorNets showed better performance than GoogleNet. In addition, the cold-started model achieves a higher mAP performance than the pre-trained model. It turned out that the property of the KITTI data set differs from the ImageNet data set in image sizes and types of object contents, so that transfer learning from the ImageNet data set is not helpful in this example. Interestingly, GoogleNet is less efficient for transfer learning than the FactorNets, and therefore showed more mAP drop than FactorNets. This inefficiency is because GoogleNet has more weight parameters to be learned than FactorNets. Thus, it makes it difficult for GoogleNet to escape the optima once learned from the ImageNet data set and to find new optima for a new KITTI data set than our FactorNets.

\begin{figure}[htbp]
\centerline{\includegraphics[width=0.95\columnwidth]{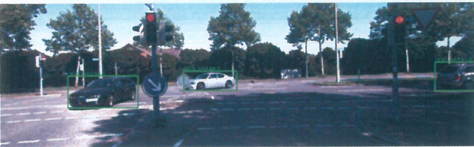}}
\caption{Example of KITTI images and car detection results by FactorNet.
	\label{fig:kitti_image}}
\end{figure}

Next, an intermediate mAPs for detecting car class during training between the proposed FactorNets and GoolgeNet was shown in Fig.~\ref{fig:map_kitti}, and an example of detection result was shown in Fig.~\ref{fig:kitti_image}. Thus, by changing GoogleNet for FactorNets in the Faster-RCNN framework, we can get \ignore{5 \%} better mAP with less computation cost. \ignore{ These improvements in mAPs of FactorNets over GoolgeNet are given in Table~\ref{tbl:detection_improvement}.}

\begin{figure}[htbp]
\centerline{\includegraphics[width=0.95\columnwidth]{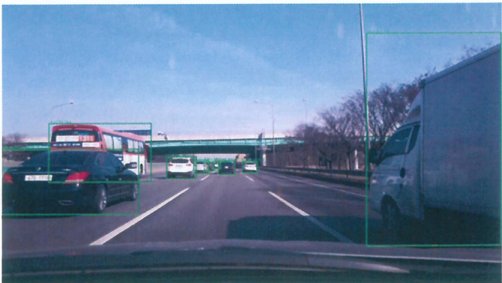}}
\caption{Example of our real-time object detection system with FactorNet.
	\label{fig:realsystem_image}}
\end{figure}

Finally, the FactorNet\_V1 in real-time object detection system in autonomous driving was incorporated to measure how it speeds up over GoogleNet in practice. Fig.~\ref{fig:realsystem_image} shows an example of 1920 $\times$ 1200 image, annotated by detection boxes inferred from FactorNet\_V1, from a sensor camera. In this real-time object inference test, FactorNet\_V1 shows 15 fps while GoogleNet results in 11 fps that is 4 fps drop in performance.

\section{Conclusions}\label{sec:conclusion}
In this study, we introduced a new lightweight CNN model, called FactorNet, that can efficiently represent deep spatial features in weight parameters and floating-point operations. Compared to GoogleNet, FactorNet model require not only 7x fewer weights but also 4x fewer floating-point operations. 

We present FactorNet\_V1 and FactorNet\_V2 as specific examples of FactorNet, and demonstrated a few practical advantages as detection accuracy and inference speed based on the KITTI data set. As a result, we are optimistic that FactorNet will be one promising lightweight, fast, and accurate CNN model for resource-restricted computing environments such as embedded devices, SOCs, and mobile phones. 


\bibliographystyle{IEEEtran}
\bibliography{ref}

\end{document}